\documentclass[11pt]{article}

\usepackage[preprint]{acl}

\usepackage{times}
\usepackage{latexsym}

\usepackage[T1]{fontenc}

\usepackage[utf8]{inputenc}

\usepackage{microtype}

\usepackage{inconsolata}

\usepackage{graphicx}
\usepackage{subfigure}
\usepackage{amsmath}
\usepackage{amssymb}
\usepackage{bm}
\usepackage{amsthm}

\usepackage{booktabs}
\usepackage{multirow}
\usepackage{xcolor}
\usepackage{colortbl}
\usepackage{subcaption}
\usepackage[ruled,vlined]{algorithm2e}

\title{SimOPD: Relational Knowledge Transfer for On-Policy Distillation}

\author{%
Kai Tang$^{12}\quad$ Xing Zheng$^2\quad$ Haobo Wang$^1$\thanks{Corresponding author.}$\quad$ Guangcheng Zhu$^{12}\quad$ Shenzhi Yang$^{12}\quad$  \\ \textbf{Zhengqing Zang$^{12}\quad$ Yingfan Ma$^2\quad$ Bowen Song$^2\quad$ Weiqiang Wang$^2\quad$} \\[4pt]
$^1$Zhejiang University \quad $^2$Ant Group \\[4pt]
\texttt{\{kai.t,wanghaobo\}@zju.edu.cn}
}

\begin{document}
\maketitle
\begin{abstract}
A key challenge in on-policy distillation (OPD) is extracting effective supervision from teacher knowledge. Supervision from token probabilities may underutilize internal knowledge, while direct hidden-state matching forces the student to shift to the teacher's feature space and complicates cross-architecture transfer. Inspired by contrastive learning's use of relationships between instances, we propose \textbf{SimOPD} to transfer teacher internal knowledge through response similarities. Specifically, SimOPD uses paired memory banks of student-generated responses to extract supervision from the teacher's internal representations as similarity distributions. 
The student then matches these distributions within its own representation space, retaining representational freedom and removing the need for cross-model projection.
Extensive experiments show that SimOPD improves the overall average over the strongest baselines by up to 2.6 and 3.1 percentage points in same- and cross-architecture settings.

\end{abstract}

\section{Introduction}
Large language models (LLMs) have achieved remarkable success~\cite{guo2025deepseek,achiam2023gpt}, demonstrating strong generalization across diverse tasks and domains~\cite{brown2020language}. To further adapt these capabilities to downstream requirements, post-training typically employs supervised fine-tuning (SFT) and reinforcement learning (RL)~\cite{ouyang2022training,chung2024scaling,guo2025deepseek}. However, SFT relies on fixed demonstrations and can suffer from training--inference mismatch~\cite{yan2026distribution}, whereas RL commonly relies on sparse outcome rewards~\cite{shao2024deepseekmath}. To address these limitations, on-policy distillation (OPD) trains on responses sampled from the student's current policy, using a frozen teacher to provide dense token-level supervision~\cite{agarwal2024policy}. This combination of on-policy learning and dense feedback has driven growing adoption of OPD in LLM post-training~\cite{yang2025qwen3,xu2026deepseek}.

\begin{figure}[t]
\centering
\includegraphics[width=0.9\linewidth]{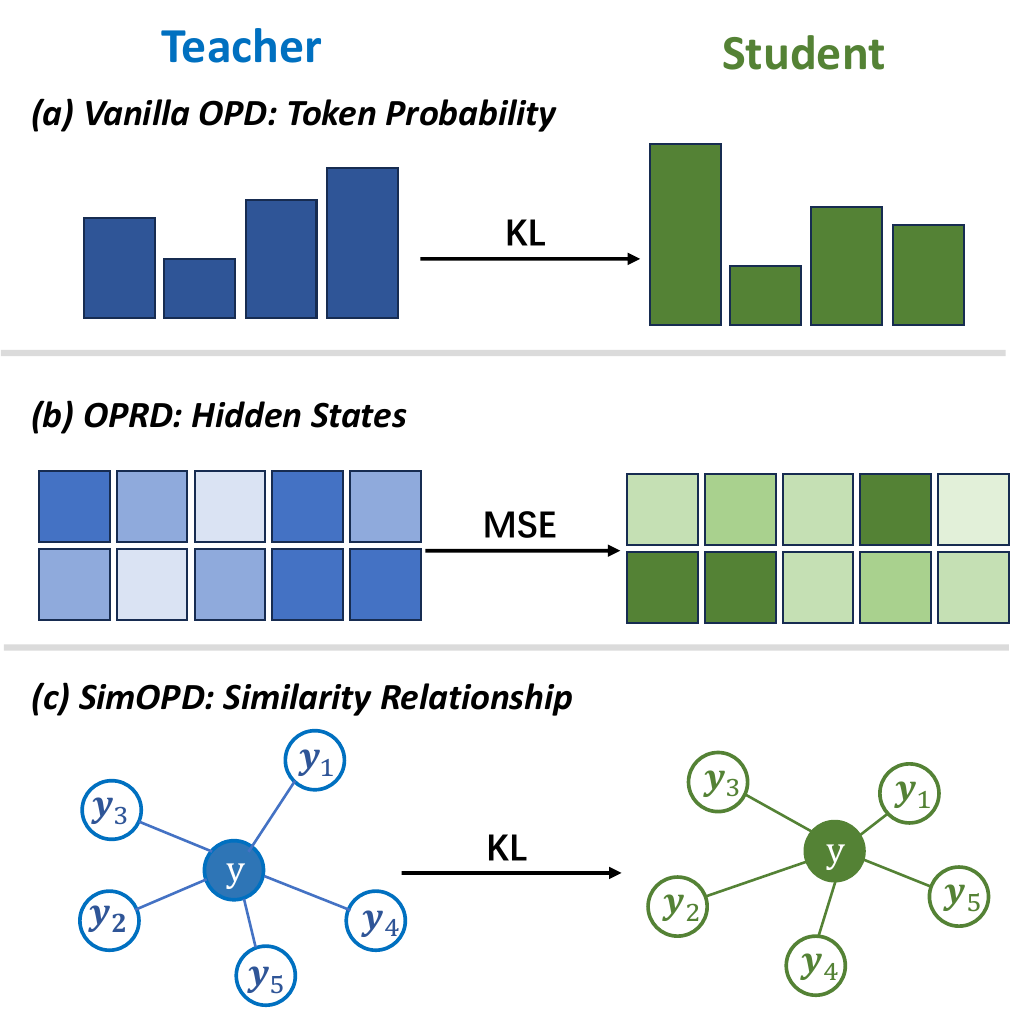}
\caption{Comparison of different on-policy distillation paradigms.}
\label{fig:intro}
\vspace{-10pt}
\end{figure}

Within this paradigm, the central question is \textit{how to extract effective supervision from the teacher’s knowledge}. Recent work has sought to improve token-level supervision. For example, ExOPD~\cite{yang2026learning} strengthens teacher rewards relative to KL regularization, PowerOPD~\cite{zhao2026poweropd} uses power-transformed probabilities to construct bounded rewards, and OPD²~\cite{heo2026policy} derives supervision from log-probability differences between a reasoning-tuned teacher and its base model. Despite these refinements, these methods still derive supervision from next-token probabilities, which can not fully convey the knowledge encoded in the teacher’s internal representations. OPRD~\cite{yang2026oprd} therefore extends on-policy distillation to hidden-state supervision, matching normalized teacher and student representations along student-generated trajectories. However, direct matching faces two limitations. First, differences in hidden dimensions and representation spaces complicate its application across architectures. OPRD-Bridge addresses this through a learned projection and a separate calibration stage. 
Second, even when representations are comparable, direct matching tightly couples the student's representations to the teacher's feature space, forcing the student to conform to the teacher's encoding and restricting its representational flexibility.
These limitations motivate a different way to \textit{extract supervision from teacher representations, without tying knowledge transfer to agreement between hidden states}.

Contrastive learning offers a useful perspective: representations can be learned through similarities and differences between instances~\cite{he2020momentum,le2020contrastive,liu2021simcls}. Inspired by this perspective, we view the teacher-defined relationships between student-generated responses as a source of supervision. Our insight is that teacher representations characterize not only individual problem–response pairs, but also how these instances relate to one another~\cite{huh2024platonic}. These relationships provide a target for knowledge transfer while allowing the student to retain its own feature encoding. In the on-policy setting, we construct this supervision over responses sampled from the student’s current policy, guiding the student to preserve the teacher-defined similarity structure within its own representation space.

To realize this idea, we propose \textbf{Sim}ilarity-based \textbf{O}n-\textbf{P}olicy \textbf{D}istillation (dubbed \textbf{SimOPD}), which transfers teacher internal knowledge through relationships between student-generated responses.
We maintain two queues: one stores teacher representations of past responses, and the other stores student representations of those same responses in the same order. For a newly generated response, each model compares its current representation with the stored representations in its own queue. This produces a similarity distribution describing how the current response relates to each past response. Since corresponding queue entries represent the same response, the teacher and student similarity distributions can be directly compared despite differences in their feature spaces. We train the student to match the teacher’s distribution and add the current representation pair to the queues for subsequent comparisons. In this way, SimOPD exploits teacher internal knowledge beyond token probabilities, while preserving the student’s representational freedom. Extensive experiments on mathematical reasoning and code generation show that our work improves overall average performance over the strongest baseline by \textbf{2.6\%} and \textbf{3.1\%}  in same- and cross-architecture settings respectively.

\begin{figure*}[!t]
  \centering
  \includegraphics[width=0.85\linewidth]{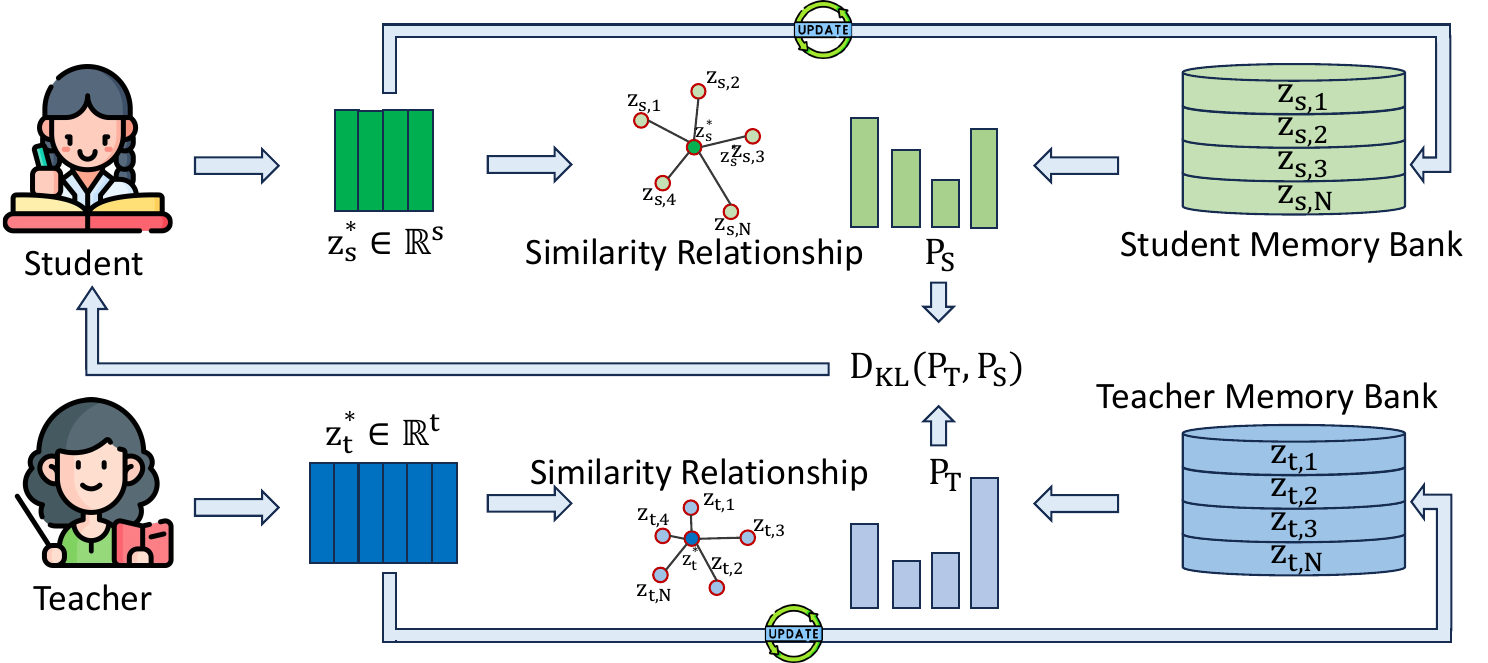}
   \caption{Overview of SimOPD. For each student-generated response, the teacher and student first compute similarity distributions over paired historical responses using their respective memory banks. The student then minimizes the KL divergence between these distributions, thereby learning teacher-defined response relationships.}
   \label{fig:method}
\end{figure*}

\section{Preliminary}
\label{sec:preliminary}

We consider a trainable student $\pi_\theta$ and a frozen teacher $\pi_T$ sharing a vocabulary $\mathcal{V}$. Given a prompt $x\sim\mathcal{D}$, the student generates a response $y=(y_1,\ldots,y_T)\sim\pi_\theta(\cdot\mid x)$. Teacher supervision is evaluated on these student-generated trajectories.

\subsection{Output-Space On-Policy Distillation}

OPD supervises the student's next-token predictions on its own generated prefixes~\cite{agarwal2024policy}. Let $p_{\theta,t}=\pi_\theta(\cdot\mid x,y_{<t})$ and $p_{T,t}=\pi_T(\cdot\mid x,y_{<t})$. The reverse-KL objective is
\begin{equation} \mathcal{L}_{\mathrm{OPD}}(\theta)=\mathbb{E}_{x,y}\left[\sum_{t=1}^{T}D_{\mathrm{KL}}\!\left(p_{\theta,t}\,\|\,p_{T,t}\right)\right], \label{eq:opd} \end{equation}
where the expectation is over prompts and student-generated responses. In an RL-style implementation, the teacher provides token-level rewards through the log-probability difference:
\begin{equation} r_t=\log\pi_T(y_t\mid x,y_{<t})-\log\pi_\theta(y_t\mid x,y_{<t}). \label{eq:opd_reward} \end{equation}
The student is trained using the corresponding policy gradient:
\begin{equation} \nabla_\theta J_{\mathrm{OPD}}(\theta)=\mathbb{E}_{x,y}\left[\sum_{t=1}^{T}r_t\nabla_\theta\log\pi_\theta(y_t\mid x,y_{<t})\right]. \label{eq:opd_gradient} \end{equation}



\subsection{On-Policy Representation Distillation}

OPRD instead supervises the student's hidden states on the same generated trajectories~\cite{yang2026oprd}. Let $h_{\theta,t}^{(\ell)}$ and $h_{T,t}^{(\ell)}$ denote comparable student and teacher representations at layer $\ell$ and response position $t$. With normalized states $\bar{h}=h/\|h\|_2$, we define the representation error at each position as
\begin{equation} \delta_t^{(\ell)}=\left\|\bar{h}_{\theta,t}^{(\ell)}-\operatorname{sg}\!\left(\bar{h}_{T,t}^{(\ell)}\right)\right\|_2^2. \label{eq:representation_error} \end{equation}
The OPRD objective averages this error over supervised positions and layers:
\begin{equation} \mathcal{L}_{\mathrm{OPRD}}(\theta)=\mathbb{E}_{x,y}\left[\frac{1}{|\mathcal{I}|}\sum_{\ell\in\mathcal{I}}\frac{\sum_t m_t\,\delta_t^{(\ell)}}{\sum_t m_t}\right], \label{eq:oprd} \end{equation}
where $\mathcal{I}$ contains the supervised layers, $m_t\in\{0,1\}$ indicates whether response position $t$ is supervised, and $\operatorname{sg}$ denotes the stop-gradient operator. When hidden spaces are incompatible, OPRD-Bridge introduces extra calibration training stage to make the representations comparable.

\section{Method}
\label{sec:method}
In this section, we introduce SimOPD, a similarity-based framework for on-policy representation distillation. As shown in Fig~\ref{fig:method}, SimOPD first constructs paired teacher and student dynamic memory banks, then transfers teacher internal knowledge by matching similarity distributions over the stored responses.

\subsection{Dual Memory Bank Construction}
\label{sec:memory_banks}

\paragraph{Response Representation Extraction.}
For a given model and a student-generated response $(x,y)$, let $\mathbf{h}_t^{(\ell)}\in\mathbb{R}^{d}$ denote the hidden state at layer $\ell$ and the $t$-th selected response position. Following Section~\ref{sec:preliminary}, we define the normalized hidden state as $\bar{\mathbf{h}}_t^{(\ell)}=\mathbf{h}_t^{(\ell)}/\|\mathbf{h}_t^{(\ell)}\|_2$. Given the selected layer set $\mathcal{I}$, the response representation is $\mathbf{z}=\frac{1}{|\mathcal{I}|T}\sum_{\ell\in\mathcal{I}}\sum_{t=1}^{T}\bar{\mathbf{h}}_t^{(\ell)}$, where $T$ is the number of selected response positions. This aggregates the normalized hidden states into a single response representation for historical comparison.

\paragraph{Dual Memory Bank Initialization.}
Using the response representations defined above, we construct a teacher memory bank $\mathcal{B}_T=(\mathbf{z}_{T,1},\ldots,\mathbf{z}_{T,N})$ and a student memory bank $\mathcal{B}_S=(\mathbf{z}_{S,1},\ldots,\mathbf{z}_{S,N})$, each with capacity $K$. The two banks store representations of the same $N\leq K$ historical responses in the same order, with $\mathbf{z}_{T,j}\in\mathbb{R}^{d_T}$ and $\mathbf{z}_{S,j}\in\mathbb{R}^{d_S}$ corresponding to response $y_j$. This correspondence provides shared response indices, while separate banks allow similarity comparisons within each model's own feature space, enabling cross-architecture distillation without cross-model projection. To initialize this shared reference set, we populate the initially empty banks using the first training batch, without updating the student model parameters.

\paragraph{Dynamic Memory Bank Updates.}
During each training update, the stored representations remain fixed. After updating the student, we append the paired representations extracted beforehand to both banks in the same order. When capacity is exceeded, the oldest pairs are removed following a first-in, first-out rule. These paired updates continually renew the shared historical references with student-generated responses for subsequent similarity-based distillation.

\subsection{Similarity-Based Distillation}
\label{sec:similarity_distillation}

\paragraph{Similarity Distributions.}
Using the dual memory banks, we characterize each new student-generated response $y$ through its similarity relationships with historical responses. Specifically, for each model $M\in\{T,S\}$, we compute the cosine similarity $\operatorname{sim}(\mathbf{z}_M^*,\mathbf{z}_{M,j})$ between its current response representation and each stored representation. We then convert these similarities into a distribution over the $N$ historical responses:
\begin{equation} P_M(j\mid y)=\frac{\exp(\operatorname{sim}(\mathbf{z}_M^*,\mathbf{z}_{M,j})/\tau)}{\sum_{r=1}^{N}\exp(\operatorname{sim}(\mathbf{z}_M^*,\mathbf{z}_{M,r})/\tau)}, \label{eq:similarity_distribution} \end{equation}
where $\tau>0$ is the temperature. Since corresponding entries in the two banks refer to the same responses, $P_T$ and $P_S$ describe relationships to a common reference set, making them directly comparable despite differences between the models' representation spaces.

\paragraph{Distillation Objective.}
To transfer these response relationships, we use the teacher's similarity distribution as a soft target and train the student by minimizing
\begin{equation} \mathcal{L}_{\mathrm{SimOPD}}(\theta)=\mathbb{E}_{x,y}\left[D_{\mathrm{KL}}\!\left(P_T(\cdot\mid y)\,\|\,P_S(\cdot\mid y)\right)\right], \label{eq:simopd_loss} \end{equation}
where the expectation is over prompts and responses sampled from the student's current policy. During optimization, the teacher and stored memory bank representations remain fixed, while gradients propagate through the current student representation. Together, the dynamic memory banks and similarity matching objective enable SimOPD to transfer teacher internal knowledge through response relationships, extending supervision beyond token probabilities while preserving the student's representational freedom.

\begin{algorithm}[t]
\footnotesize
\caption{SimOPD}
\label{alg:simopd}
\KwIn{Prompt dataset $\mathcal{D}$, student $\pi_\theta$, frozen teacher $\pi_T$, training steps $U$, bank capacity $K$, temperature $\tau$.}
\KwOut{Optimized student $\pi_\theta$.}

Initialize dual memory banks $\mathcal{B}_T,\mathcal{B}_S\leftarrow\emptyset$\;

\For{$u=1,\ldots,U$}{
    \tcp{On-policy response representations}
    Sample prompts from $\mathcal{D}$ and generate a response batch $\mathcal{R}=\{(x_i,y_i)\}_{i=1}^{B}$ using $\pi_\theta$\;
    Extract teacher and student response representations $\{\mathbf{z}_{T,i},\mathbf{z}_{S,i}\}_{i=1}^{B}$\;

    \tcp{Similarity-based distillation}
    $N\leftarrow|\mathcal{B}_T|=|\mathcal{B}_S|$\;
    \If{$N\geq2$}{
        \For{$i=1,\ldots,B$}{
            Compute for $M\in\{T,S\}$ and $j=1,\ldots,N$:\;
            $s_{M,ij}\leftarrow\cos(\mathbf{z}_{M,i},\mathcal{B}_M[j])/\tau$\;
            $\mathbf{p}_{M,i}\leftarrow\operatorname{softmax}((s_{M,ij})_{j=1}^{N})$\;
        }
        $\mathcal{L}\leftarrow\frac{1}{B}\sum_{i=1}^{B}D_{\mathrm{KL}}(\mathbf{p}_{T,i}\,\|\,\mathbf{p}_{S,i})$\;
        Update $\theta$ by minimizing $\mathcal{L}$\;
    }
    Dual memory bank update;
}
\Return{$\pi_\theta$}\;
\end{algorithm}

\section{Experiment}
\label{sec:experiment}
\subsection{Setup}
\paragraph{Implementation Details.}
Our implementation builds on verl~\cite{verl} with vLLM for on-policy generation. We consider Qwen3-4B $\rightarrow$ Qwen3-4B-Base for same-architecture distillation and Qwen3-8B $\rightarrow$ Qwen3-1.7B-Base for cross-architecture distillation~\cite{yang2025qwen3}. For mathematical reasoning and code generation, we train on DAPO-Math-17K~\cite{yu2026dapo} and deduplicated coding subset of Eurus~\cite{cui2025process,yuan2024implicitprm}, respectively.  For SimOPD, we extract final-layer representations from up to the last 2,000 response tokens, set each memory bank's capacity to $K=2048$, and use a shared similarity temperature of $\tau=0.1$. Further implementation details are provided in the appendix.

\paragraph{Evaluation.}
Following OPRD~\cite{yang2026oprd}, we evaluate mathematical reasoning on AIME 2024 and AIME 2025, and AIMO (AMC 2022/2023)~\cite{li2024numinamath}, reporting Avg@16 over 16 independently sampled responses per problem at temperature $0.7$. For code generation, we conduct evaluation on HumanEval~\cite{chen2021evaluating}, MBPP~\cite{austin2021program}, and LeetCode~\cite{guo2024deepseekcoder}, reporting pass@1 based on benchmark test execution. Teachers, initial students, and distilled models use consistent decoding settings within each benchmark.

\paragraph{Baselines.}
We compare SimOPD with six distillation baselines: (1) \textbf{Sampled-token OPD}~\cite{agarwal2024policy} uses teacher--student log-probability differences on student-sampled tokens. (2) \textbf{Top-$k$ OPD}~\cite{agarwal2024policy} computes the distillation loss over the student's top-$k$ tokens and we take $k$ as 16 in experiments. (3) \textbf{ExOPD}~\cite{yang2026learning} strengthens teacher-derived rewards relative to reference-policy KL regularization. (4) \textbf{PowerOPD}~\cite{zhao2026poweropd} constructs bounded rewards through power transformations of teacher and student probabilities. (5) \textbf{OPD$^2$}~\cite{heo2026policy} derives supervision from log-probability differences between the teacher and its pretrained base model, with reward centering and sign-consistency gating. (6) \textbf{OPRD}~\cite{yang2026oprd} matches normalized teacher and student hidden states along student-generated trajectories, using \textbf{OPRD-Bridge} in the cross-architecture setting. Additionally, for representation-based methods OPRD and SimOPD, we also evaluate joint training with sampled-token OPD loss, setting the OPD loss coefficient to $1$.

\subsection{Main Result}

\begin{table*}[t]
\centering
\small
\setlength{\tabcolsep}{3pt}
\renewcommand{\arraystretch}{1}
\caption{Same-architecture results: Qwen3-4B $\rightarrow$ Qwen3-4B-Base. Math and code use separate training runs. Scores are percentages; Avg. is the mean within each domain, and Overall averages the two domains before rounding. Indented + OPD denotes joint training. Bold and underlined numbers mark the best and second-best distilled results, respectively.}
\label{tab:same_arch_results}
\resizebox{\textwidth}{!}{%
\begin{tabular}{l*{9}{c}}
\toprule
\multirow{2}{*}{\textbf{Method}} & \multicolumn{4}{c}{\textbf{Mathematical Reasoning}} & \multicolumn{4}{c}{\textbf{Code Generation}} & \multirow{2}{*}{\textbf{Overall}} \\
\cmidrule(lr){2-5}\cmidrule(lr){6-9}
 & AIME24 & AIME25 & AIMO & Avg. & HumanEval & LeetCode & MBPP & Avg. & \\
\midrule
Teacher & 25.8 & 17.5 & 59.0 & 34.1 & 87.8 & 73.3 & 70.4 & 77.2 & 55.6 \\
Initial Student & 7.4 & 5.1 & 27.1 & 13.2 & 57.9 & 43.2 & 40.4 & 47.2 & 30.2 \\
\midrule
OPD (sampled-token) & 9.4 & 8.3 & 37.2 & 18.3 & 77.2 & 40.4 & \textbf{71.2} & 62.9 & 40.6 \\
OPD (top-$k$) & 9.0 & 8.5 & 35.9 & 17.8 & 75.8 & 40.5 & 67.1 & 61.1 & 39.5 \\
ExOPD & 11.7 & 13.3 & 42.2 & 22.4 & 81.1 & 41.7 & 67.7 & 63.5 & 43.0 \\
PowerOPD & \textbf{16.7} & 11.7 & \textbf{44.9} & 24.4 & 72.5 & 33.9 & 64.9 & 57.1 & 40.8 \\
OPD$^2$ & 12.2 & 14.2 & 42.3 & 22.9 & 78.0 & 35.6 & 67.2 & 60.3 & 41.6 \\
OPRD & 12.5 & 7.5 & 40.9 & 20.3 & 79.3 & 37.2 & 62.2 & 59.6 & 39.9 \\
\quad \textit{+ OPD} & 13.1 & 12.5 & 40.9 & 22.2 & 79.7 & 41.9 & 67.3 & 63.0 & 42.6 \\
\midrule
\rowcolor{gray!12}
\textbf{SimOPD (ours)} & 15.6 & \underline{14.8} & 43.7 & \underline{24.7} & \textbf{84.2} & \underline{43.9} & 67.9 & \underline{65.3} & \underline{45.0} \\
\rowcolor{gray!12}
\quad \textit{+ OPD} & \underline{15.8} & \textbf{15.3} & \underline{44.1} & \textbf{25.1} & \underline{83.9} & \textbf{44.2} & \underline{70.2} & \textbf{66.1} & \textbf{45.6} \\
\bottomrule
\end{tabular}%
}
\end{table*}

\begin{table*}[t]
\centering
\small
\setlength{\tabcolsep}{3pt}
\renewcommand{\arraystretch}{1}
\caption{Cross-architecture results: Qwen3-8B $\rightarrow$ Qwen3-1.7B-Base. Math and code use separate training runs. Scores are percentages; Avg. is the mean within each domain, and Overall averages the two domains before rounding. Indented + OPD denotes joint training. Bold and underlined numbers mark the best and second-best distilled results, respectively.}
\label{tab:cross_arch_results}
\resizebox{\textwidth}{!}{%
\begin{tabular}{l*{9}{c}}
\toprule
\multirow{2}{*}{\textbf{Method}} & \multicolumn{4}{c}{\textbf{Mathematical Reasoning}} & \multicolumn{4}{c}{\textbf{Code Generation}} & \multirow{2}{*}{\textbf{Overall}} \\
\cmidrule(lr){2-5}\cmidrule(lr){6-9}
 & AIME24 & AIME25 & AIMO & Avg. & HumanEval & LeetCode & MBPP & Avg. & \\
\midrule
Teacher & 25.8 & 20.8 & 58.4 & 35.0 & 88.4 & 68.9 & 70.9 & 76.1 & 55.5 \\
Initial Student & 0.8 & 0.8 & 11.7 & 4.4 & 61.4 & 21.8 & 52.2 & 45.1 & 24.8 \\
\midrule
OPD (sampled-token) & 4.2 & 3.7 & 23.5 & 10.5 & 64.0 & 23.3 & 53.4 & 46.9 & 28.7 \\
OPD (top-$k$) & 5.0 & 4.3 & 22.9 & 10.7 & 61.9 & 22.0 & 53.5 & 45.8 & 28.3 \\
ExOPD & 7.5 & 4.2 & 22.8 & 11.5 & 62.8 & 21.7 & 53.1 & 45.9 & 28.7 \\
PowerOPD & \underline{11.2} & 5.6 & 33.8 & 16.9 & 63.4 & 18.9 & 56.4 & 46.2 & 31.6 \\
OPD$^2$ & 6.7 & 1.7 & 22.9 & 10.4 & 69.5 & \underline{26.8} & \textbf{57.4} & \underline{51.2} & 30.8 \\
OPRD-Bridge & 8.8 & \textbf{8.3} & 33.6 & 16.9 & 63.8 & 21.7 & 52.4 & 46.0 & 31.4 \\
\quad \textit{+ OPD} & 9.1 & \underline{8.0} & 33.4 & 16.8 & 66.2 & 21.8 & 54.9 & 47.6 & 32.2 \\
\midrule
\rowcolor{gray!12}
\textbf{SimOPD (ours)} & 10.7 & \textbf{8.3} & \underline{34.6} & \underline{17.9} & \textbf{72.6} & 26.7 & 49.1 & 49.5 & \underline{33.7} \\
\rowcolor{gray!12}
\quad \textit{+ OPD} & \textbf{11.9} & \textbf{8.3} & \textbf{35.5} & \textbf{18.6} & \underline{72.5} & \textbf{27.1} & \underline{56.6} & \textbf{52.1} & \textbf{35.3} \\
\bottomrule
\end{tabular}%
}
\end{table*}

We evaluate SimOPD on mathematical reasoning and code generation under same- and cross-architecture distillation, with separate training runs for each domain. Tables~\ref{tab:same_arch_results} and~\ref{tab:cross_arch_results} show that SimOPD ranks among the top two standalone methods in all four settings, leads in three, and achieves the highest Overall score in both model pairs. Moreover, adding sampled-token OPD yields the highest domain averages in all four settings.

\paragraph{Same-Architecture Results.}
Relative to the strongest baselines, PowerOPD and ExOPD, SimOPD improves mathematical reasoning and code generation averages by 0.3 and 1.8 percentage points, reaching 24.7 and 65.3, respectively.
Although different baselines lead in the two domains, SimOPD performs strongly across both, achieving the highest standalone Overall score of 45.0 and exceeding ExOPD by 2.0 points. Notably, SimOPD also surpasses OPRD by 4.4 and 5.7 points in mathematics and code generation, respectively, demonstrating the advantage of matching response similarities over directly matching hidden states. Building on these gains, adding sampled-token OPD raises the domain averages to 25.1 and 66.1 and the Overall score to 45.6, exceeding the strongest baseline by 2.6 points. In code generation, this additional supervision particularly improves MBPP, from 67.9 to 70.2, while maintaining strong HumanEval and LeetCode performance. The resulting improvements across both domains support combining similarity-based supervision with direct token-level feedback.

\paragraph{Cross-Architecture Results.}
SimOPD improves the mathematical reasoning average by 1.0 percentage point over the strongest baselines, PowerOPD and OPRD-Bridge, reaching 17.9 and ranking first among standalone methods.
Meanwhile, it ranks second among standalone methods in code generation and achieves a HumanEval score of 72.6, surpassing the strongest baseline, OPD$^2$, by 3.1 points. Together, these gains yield the highest standalone Overall score of 33.7, exceeding OPRD-Bridge+OPD by 1.5 points. These results support the effectiveness of transferring teacher internal knowledge across different model capacities through similarities computed within each model's own representation space. However, SimOPD shows a performance drop on MBPP, which we attribute to its relatively simple tasks and sensitivity to output format. We provide a detailed analysis in Appendix~\ref{app:mbpp_analysis}. Furthermore, adding OPD raises the mathematical reasoning and code generation averages to 18.6 and 52.1, outperforming the strongest baselines by 1.7 and 0.9 points, respectively. Consequently, the Overall score reaches 35.3, extending the advantage over the strongest baseline to 3.1 points. Relative to standalone SimOPD, joint training improves code generation by 2.6 points and mathematics by 0.7 points, suggesting that direct token-level feedback provides particularly useful complementary supervision for this smaller student's code generation.

These results support relational knowledge transfer across architectures and its complementarity with token-level supervision.

\subsection{Further Analysis}

\subsubsection{Ablation Study}
We examine three SimOPD variants to assess its memory banks and distillation objective. \textbf{In-Batch Matching} replaces historical banks with paired teacher and student representations from the current batch, retaining the KL objective. \textbf{Similarity Regression} retains the dual banks but applies MSE to teacher and student cosine similarities to corresponding historical responses. \textbf{Single Memory Bank} uses the teacher bank as a shared reference for both models. In cross-architecture settings, an MLP maps student queries to the teacher's representation dimension.
Table~\ref{tab:math_ablations} supports the benefits of historical references and distribution matching. 
Historical memory banks improve the average score over In-Batch Matching by 1.9 and 5.5 percentage points in the same- and cross-architecture settings, respectively, highlighting the value of historical comparisons, particularly for heterogeneous models.
Meanwhile, Similarity Regression underperforms SimOPD by 1.5 and 0.7 points, favoring normalized similarity distributions over direct regression of cosine values. Finally, Single Memory Bank lowers the same-architecture average to 22.8, indicating that separate reference spaces remain beneficial even when hidden dimensions match. Together, these results support matching response relationships through paired historical representations rather than a shared feature space or raw similarity scores.

\begin{table}[t]
\centering
\small
\setlength{\tabcolsep}{2.5pt}
\renewcommand{\arraystretch}{1.3}
\caption{Ablation study of SimOPD.}
\label{tab:math_ablations}

\textbf{(a) Same-architecture}\par
Qwen3-4B $\rightarrow$ Qwen3-4B-Base\par
\smallskip

\begin{tabular}{lcccc}
\toprule
Method & AIME24 & AIME25 & AIMO & Avg. \\
\midrule
\rowcolor{gray!15}
\textbf{SimOPD} & 15.6 & 14.8 & 43.7 & 24.7\\
In-Batch Matching & 14.7 & 12.5 & 41.3 & 22.8 \\
Similarity Regression & 15.2 & 11.9 & 42.5 & 23.2 \\
Single Memory Bank & 13.6 & 12.7 & 42.2 & 22.8 \\
\bottomrule
\end{tabular}

\par\vspace{14pt}

\textbf{(b) Cross-architecture}\par
Qwen3-8B $\rightarrow$ Qwen3-1.7B-Base\par
\smallskip

\begin{tabular}{lcccc}
\toprule
Method & AIME24 & AIME25 & AIMO & Avg. \\
\midrule
\rowcolor{gray!15}
\textbf{SimOPD} & 10.7 & 8.3 & 34.6 & 17.9 \\
In-Batch Matching & 5.0 & 3.3 & 28.8 & 12.4 \\
Similarity Regression &9.8  & 8.1 & 33.6  &17.2  \\
Single Memory Bank & 0.3 & 0.3  & 9.2 & 3.3  \\
\bottomrule
\end{tabular}
\vspace{-10pt}
\end{table}

\subsubsection{Sensitivity Analysis}
\begin{figure*}[!t]
  \centering
  \includegraphics[width=0.9\linewidth]{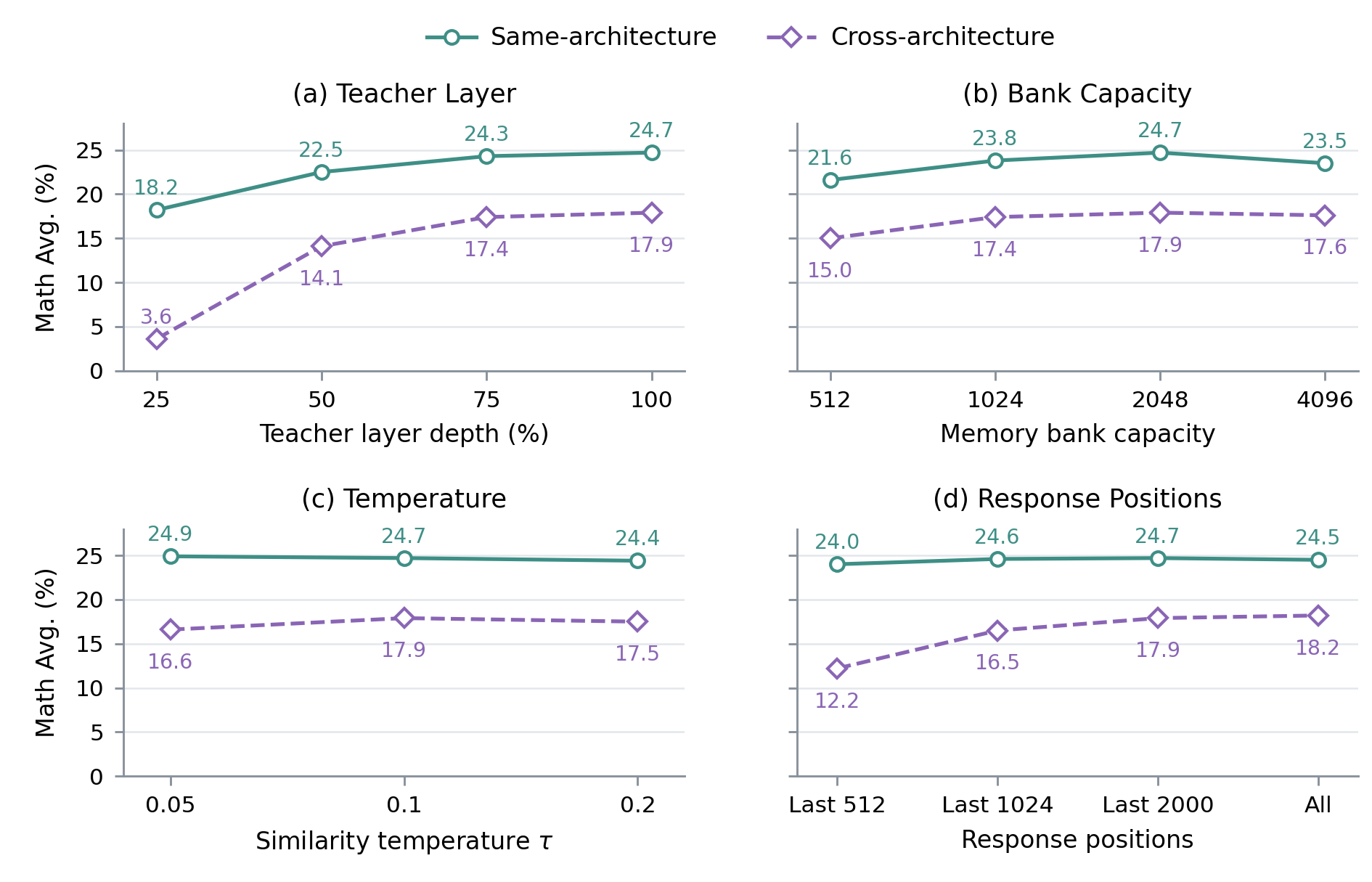}
   \caption{Sensitivity of SimOPD to (a) teacher layer depth, (b) memory bank capacity, (c) similarity temperature, and (d) response positions. Math Avg. denotes the mean Avg@16 (\%) across AIME24, AIME25, and AIMO. Each panel varies one factor while keeping the remaining settings fixed.}
   \label{fig:sensitivity}
\end{figure*}

We examine four factors in both same- and cross-architecture mathematical reasoning settings, varying one factor at a time under the same training budget and evaluation protocol. We select teacher layers at 25\%, 50\%, 75\%, and 100\% depth while retaining the student's final layer. We also vary bank capacity $K\in\{512,1024,2048,4096\}$, similarity temperature $\tau\in\{0.05,0.1,0.2\}$, and response coverage over the last 512, 1,024, and 2,000 tokens or all response tokens.
\vspace{-5pt}
\paragraph{Teacher Layer Selection.}
Figure~\ref{fig:sensitivity} shows that deeper teacher layers consistently improve performance, with diminishing gains near the final layer. Moving from 25\% to 100\% depth improves the average by 6.5 and 14.3 percentage points in the same- and cross-architecture settings, respectively. This suggests that useful teacher internal knowledge is more effectively exposed by later-layer representations, particularly when teacher and student differ in capacity.
\vspace{-5pt}
\paragraph{Memory Bank Capacity.}
SimOPD is relatively insensitive to the tested memory bank capacities with $K\geq1024$, maintaining strong performance with a compact historical reference set. 
With only half the reference-cache storage, a bank of 1,024 remains within 0.9 and 0.5 percentage points of the 2,048 baseline in the same- and cross-architecture settings, respectively.
Moreover, increasing the capacity to 4,096 yields no further performance gains in either setting. These results suggest that transferring teacher internal knowledge does not require an extensive reference cache, offering promising potential for larger-scale applications without proportional growth in cache storage.
\vspace{-5pt}
\paragraph{Similarity Temperature.}
SimOPD exhibits strong robustness across the tested temperatures, with average scores varying by only 0.5 and 1.3 percentage points in the same- and cross-architecture settings, respectively. Although temperature changes the sharpness of the similarity distributions, effective transfer of teacher internal knowledge is maintained across these choices. This stability across both model pairs provides flexibility in temperature selection, suggesting limited dependence on precise tuning within the tested range.
\vspace{-5pt}
\paragraph{Response Coverage.}
Expanding coverage from the last 512 to the last 2,000 tokens improves performance by 0.7 and 5.7 points, respectively, indicating greater sensitivity in cross-architecture distillation. However, the last 2,000 tokens remain within 0.3 points of full-response pooling in both settings. Since later hidden states condition on preceding context, they can retain useful information beyond the selected suffix, helping explain why this coverage captures sufficient teacher internal knowledge for effective distillation.

\subsubsection{Out-of-Domain Generalization}
To assess generalization beyond the training domain, we evaluate checkpoints obtained after mathematical reasoning distillation on ARC-C~\cite{clark2018think}, GPQA~\cite{rein2023gpqa}, and MMLU-Pro~\cite{wang2024mmlu}. These benchmarks cover school-level science questions, graduate-level scientific reasoning, and multidisciplinary knowledge and reasoning, respectively.
\begin{table}[t]
\centering
\small
\setlength{\tabcolsep}{4pt}
\renewcommand{\arraystretch}{1.10}
\caption{OOD evaluation with Qwen3-4B-Base.}
\label{tab:ood_results}
\resizebox{\linewidth}{!}{%
\begin{tabular}{lcccc}
\toprule
Method & ARC-C & GPQA & MMLU-Pro & Avg. \\
\midrule
Initial Student & 31.0 & 14.1 & 32.5 & 25.9 \\
\midrule
OPD & 67.6 & 21.5 & 43.5 & 44.2 \\
PowerOPD & 68.2 & 23.4 & 51.3 & 47.6 \\
OPD$^2$ & 61.7 & 21.8 & 44.9 & 42.8 \\
\midrule
\rowcolor{gray!12}
\textbf{SimOPD (ours)} & 72.3 & 23.4 & 50.7 & 48.8 \\
\bottomrule
\end{tabular}%
}
\vspace{-20pt}
\end{table}
As shown in Table~\ref{tab:ood_results}, all evaluated distillation methods improve upon the initial student across the three benchmarks. SimOPD raises the average from 25.9 to 48.8, showing that its gains on the target tasks are accompanied by improved performance on these OOD benchmarks. Moreover, it exceeds the strongest baseline, PowerOPD, by 1.2 percentage points on average, gaining 4.1 points on ARC-C and matching its GPQA score, while trailing by 0.6 points on MMLU-Pro. Although these benchmarks are not directly optimized during training, SimOPD achieves the highest OOD average, supporting the broader generalization benefits of similarity-based knowledge transfer.

\section{Related Work}

\subsection{On-Policy Distillation}
On-policy distillation learns from student-generated responses using teacher feedback. GKD explores different divergence objectives on these responses~\cite{agarwal2024policy}, while MiniLLM optimizes reverse KL through policy gradients~\cite{gu2024minillm}. Building on this paradigm, recent methods refine the token-level supervision: ExOPD strengthens teacher rewards relative to KL regularization~\cite{yang2026learning}, PowerOPD constructs bounded rewards through probability transformations~\cite{zhao2026poweropd}, and OPD$^2$ isolates reasoning-related supervision through log-probability differences between a tuned teacher and its base model~\cite{heo2026policy}. Despite these different constructions, their supervision originates from next-token probabilities.
OPRD extends this paradigm to internal representations by matching normalized teacher and student hidden states along student-generated trajectories~\cite{yang2026oprd}. For heterogeneous models, OPRD-Bridge introduces calibrated projections to enable representation matching. SimOPD likewise extracts supervision from teacher representations, but transfers similarities between responses through paired memory banks. 
By comparing responses within each model's own representation space, SimOPD removes the need for a learned cross-model projection.
\vspace{-5pt}

\subsection{Representations Research in LLMs}
Our approach draws inspiration from well-established methods in representation learning, including Similarity-Preserving Knowledge Distillation and CompRess~\cite{tung2019similarity,abbasi2020compress}.
Similarly, a growing body of work adapts established representation-based techniques to LLM learning and assessment. Minitron investigates intermediate-state distillation for recovering pruned models~\cite{muralidharan2024compact}. In RLVR, Hidden-Align aligns representations of correct rollouts~\cite{wang2026right}, while SHEAR uses hidden-state distributional differences for credit assignment~\cite{chen2026hidden}. Meanwhile, GeoMin uses representation geometry to assess pseudo-label reliability and select training instances~\cite{zhu2026geomin}, and INSIDE detects hallucinations through response-embedding covariance spectra~\cite{chen2024inside}. Within this line of research, SimOPD applies similarity-based distillation to on-policy language generation, transferring teacher internal knowledge through similarities between student-generated responses.

\section{Conclusion}
We introduced SimOPD, an on-policy distillation method that transfers teacher internal knowledge through similarities between student-generated responses. Its paired memory banks enable similarity-distribution matching within each model's own representation space, preserving the student's representational freedom and supporting different hidden dimensions without cross-model projection. Experiments on mathematical reasoning and code generation demonstrate strong performance in both same- and cross-architecture settings, with further gains from combining SimOPD with token-level OPD. Meanwhile, ablation studies support the memory bank design and distillation objective. These findings support response similarities as an effective source of supervision for utilizing teacher internal knowledge beyond token probabilities.

\section*{Limitations}
Our evaluation primarily covers reasoning tasks, including mathematical reasoning and code generation. Applicability to open-ended generation and other modalities remains to be investigated.
Compared with sampled-token supervision, representation-based relational distillation requires additional hidden-state processing and cross-response similarity computation. This structural overhead limits training efficiency in our current implementation, motivating further optimization of representation extraction and relation construction. Memory bank management also offers opportunities for exploration, including alternative initialization schemes, memory warm-up schedules, and update strategies. Finally, the role of additional regularization in representation learning and similarity matching has not been systematically examined.



\bibliography{custom}

\newpage
\appendix

\section{Additional Implementation Details}
\label{app:implementation}
Across both domains, we train for 200 steps on eight NVIDIA A100 80GB GPUs using AdamW~\cite{loshchilov2017decoupled} with a learning rate of $10^{-5}$, a global batch size of 16 responses and an actor micro-batch size of 1 per GPU. Prompts longer than 3,072 tokens are filtered before training, and rollout generation uses top-$p=1.0$. Each batch contains eight prompts with two responses sampled per prompt at temperature $1.0$, using non-thinking prompting and a maximum response length of 4,096 tokens.

About baseline configurations, we follow the original formulations and model-role definitions of the baselines, adapting them to our common teacher--student pairs. ExOPD uses a reward weight of $\lambda=1.25$ and the initial student as its reference model, while PowerOPD uses a power exponent of $\alpha=5.0$. For OPD$^2$, we use the teacher's corresponding base model, top-1024 student tokens for reward centering, and a reward scale of $0.1$. OPRD supervises all layers over the last 2,000 response tokens, and OPRD-Bridge uses a rank-8 projection. For a controlled comparison, all methods are trained for 200 steps without learning-rate warm-up.

\section{MBPP Error Analysis}
\label{app:mbpp_analysis}

We partition incorrect MBPP responses into format errors and execution failures, with the latter covering the remaining incorrect responses after excluding format errors.

\begin{table}[t]
\centering
\small
\setlength{\tabcolsep}{4pt}
\renewcommand{\arraystretch}{1.12}
\caption{MBPP error breakdown under cross-architecture distillation. All values are percentages of evaluated samples.}
\label{tab:mbpp_errors}
\resizebox{\linewidth}{!}{%
\begin{tabular}{lccc}
\toprule
Method & Accuracy & \shortstack{Format\\errors} & \shortstack{Execution\\failures} \\
\midrule
OPD & 53.40 & 14.50 & 32.10 \\
ExOPD & 53.10 & 15.30 & 31.60 \\
PowerOPD & 56.40 & 2.01 & 41.59 \\
OPD$^2$ & 57.40 & 1.75 & 40.85 \\
\midrule
\rowcolor{gray!12}
SimOPD & 49.10 & 23.60 & 27.30 \\
\rowcolor{gray!12}
\quad \textit{+ OPD} & 56.60 & 13.20 & 30.20 \\
\bottomrule
\end{tabular}%
}
\end{table}

Table~\ref{tab:mbpp_errors} identifies format compliance as a key bottleneck for SimOPD on MBPP. Compared with OPD, SimOPD incurs 9.1 percentage points more format errors but 4.8 points fewer execution failures, resulting in a 4.3-point accuracy deficit. Adding OPD substantially reduces format errors from 23.6\% to 13.2\% and raises accuracy by 7.5 points. Together, these results suggest that direct token-level supervision has a substantial advantage over indirect representation supervision in learning the output conventions required by MBPP, motivating their joint use.

\section{Training Efficiency.}
We assess training efficiency on mathematical reasoning under same-architecture distillation from Qwen3-4B to Qwen3-4B-Base, using eight NVIDIA A100 80GB GPUs. We report average wall-clock time per training step and peak GPU memory per device, taking the maximum peak across the eight GPUs. All methods use matched hardware, batch sizes, and response-length limits. Measurements exclude the first five warm-up steps, evaluation, and checkpoint saving.

\begin{table}[t]
\centering
\small
\setlength{\tabcolsep}{5pt}
\renewcommand{\arraystretch}{1.12}
\caption{Training costs for same-architecture mathematical reasoning on eight A100 80GB GPUs. Memory denotes the maximum per-GPU peak. Lower is better.}
\label{tab:training_efficiency}
\resizebox{\linewidth}{!}{%
\begin{tabular}{lcc}
\toprule
Method & \shortstack{Time/step\\(s) $\downarrow$} & \shortstack{Peak memory\\(GB/GPU) $\downarrow$} \\
\midrule
ExOPD & 70.25 & 76.45 \\
PowerOPD & 66.92 & 76.39 \\
OPD$^2$ & 73.69 & 79.69 \\
OPRD & 83.04 & 77.56 \\
\midrule
\rowcolor{gray!12}
\textbf{SimOPD (ours)} & 87.20 & 78.77 \\
\bottomrule
\end{tabular}%
}
\end{table}

As shown in Table~\ref{tab:training_efficiency}, SimOPD improves the mathematical reasoning average over the representation-based OPRD baseline by 4.4 percentage points, with 87.20 seconds per step and a 5.0\% increase in training time. Meanwhile, its peak memory of 78.77 GB per GPU remains within the baseline range of 76.39--79.69 GB. Relative to direct hidden-state matching, SimOPD therefore achieves stronger distillation performance with a modest increase in training cost. Moreover, memory banks are used only during training, leaving the student's architecture and inference procedure uninfluenced.

\end{document}